\documentclass[10pt, a4paper]{article}

\usepackage{lrec2026} 
\usepackage{booktabs}
\usepackage[table]{xcolor} 
\usepackage{multicol}
\usepackage{subcaption}
\usepackage{amssymb}
\usepackage{microtype}
\usepackage{comment}

\definecolor{darkgreen}{HTML}{00AA00} 
\newcommand{\E}{\,\textsuperscript{\textcolor{blue}{$\bigstar$}}}
\newcommand{\SY}{\,\textsuperscript{\textcolor{orange}{$\blacksquare$}}}
\newcommand{\LI}{\,\textsuperscript{\textcolor{darkgreen}{$\blacktriangle$}}}
\newcommand{\AD}{\,\textsuperscript{\textcolor{red}{$\blacklozenge$}}}

\usepackage{paralist}
\usepackage{enumitem}
\title{\mbox{PETra: A Multilingual Corpus of Pragmatic Explicitation in Translation}}

\name{
\begin{tabular}{c}
Doreen Osmelak\textsuperscript{1}, Koel Dutta Chowdhury\textsuperscript{1}, \\ Uliana Sentsova\textsuperscript{1},
Cristina España-Bonet\textsuperscript{2,3}, Josef van Genabith\textsuperscript{1,2}
\end{tabular}
\thanks{Correspondence to: \href{mailto:dosmelak@lst.uni-saarland.de,koeldc@lst.uni-saarland.de}{{dosmelak, koeldc}@lst.uni-saarland.de}
The corpus and code can be found at \url{https://huggingface.co/datasets/Doosme/PETra} and \url{https://github.com/Doosme/PETra}
}
}

\address{\textsuperscript{1} Saarland University, Saarland Informatics Campus, Germany\\
\textsuperscript{2} German Research Center for Artificial Intelligence (DFKI)\\
\textsuperscript{3} Barcelona Supercomputing Center (BSC-CNS), Barcelona, Catalonia, Spain
\\
 }

\abstract{
Translators often enrich texts with background details that make implicit cultural meanings explicit for new audiences. This phenomenon, known as pragmatic explicitation, has been widely discussed in translation theory but rarely modeled computationally. We introduce PETra, the first multilingual corpus and detection framework for pragmatic explicitation. The corpus consists of 3,000 sentence pairs from TED-Multi and Europarl, covers twelve language pairs, and includes additions such as entity descriptions, measurement conversions, and translator remarks. We identify candidates through null alignments and refine them using active learning with human annotation. Our results show that entity and system-level (e.g., metric conversions) explicitations are most frequent, and that active learning improves classifier accuracy by $7$-$8$ percentage points, achieving up to $0.88$ accuracy and $0.82$ F1 for the best transfer languages. PETra establishes pragmatic explicitation as a measurable, cross-linguistic phenomenon and takes a step towards building culturally aware machine translation.\\\newline
}

\begin{document}

\maketitleabstract

\section{Introduction}
When translating between languages, what is left unsaid in one culture may need to be spelled out in another. Consider an English text that refers simply to \textit{Angela Merkel}. For a German audience, the name alone suffices. Yet a translator into, say, Arabic might render it as \textit{Angela Merkel, former German Chancellor}, adding background knowledge that the target audience may lack. This added phrase does not alter the literal meaning but enriches the translation by making implicit, tacit knowledge in the source explicit in the target.

This process, known as pragmatic explicitation, reflects how translators bridge gaps in shared world knowledge between source and target audiences. The source community may possess contextual or cultural knowledge that is assumed but unstated, while the target audience lacks access to such background. Translators thus introduce minimal but informative cues, such as titles, descriptions, or references, to clarify meaning and preserve communicative intent. Although explicitation has long been recognized in translation studies \citep{vinay1958stylistique, nida1964toward, klaudy1993explicitation, snell2006turns}, it remains underexplored in computational research, particularly in terms of pragmatic or culturally motivated explicitations.

\begin{figure*}[t]
    \centering
\includegraphics[width=\linewidth]
{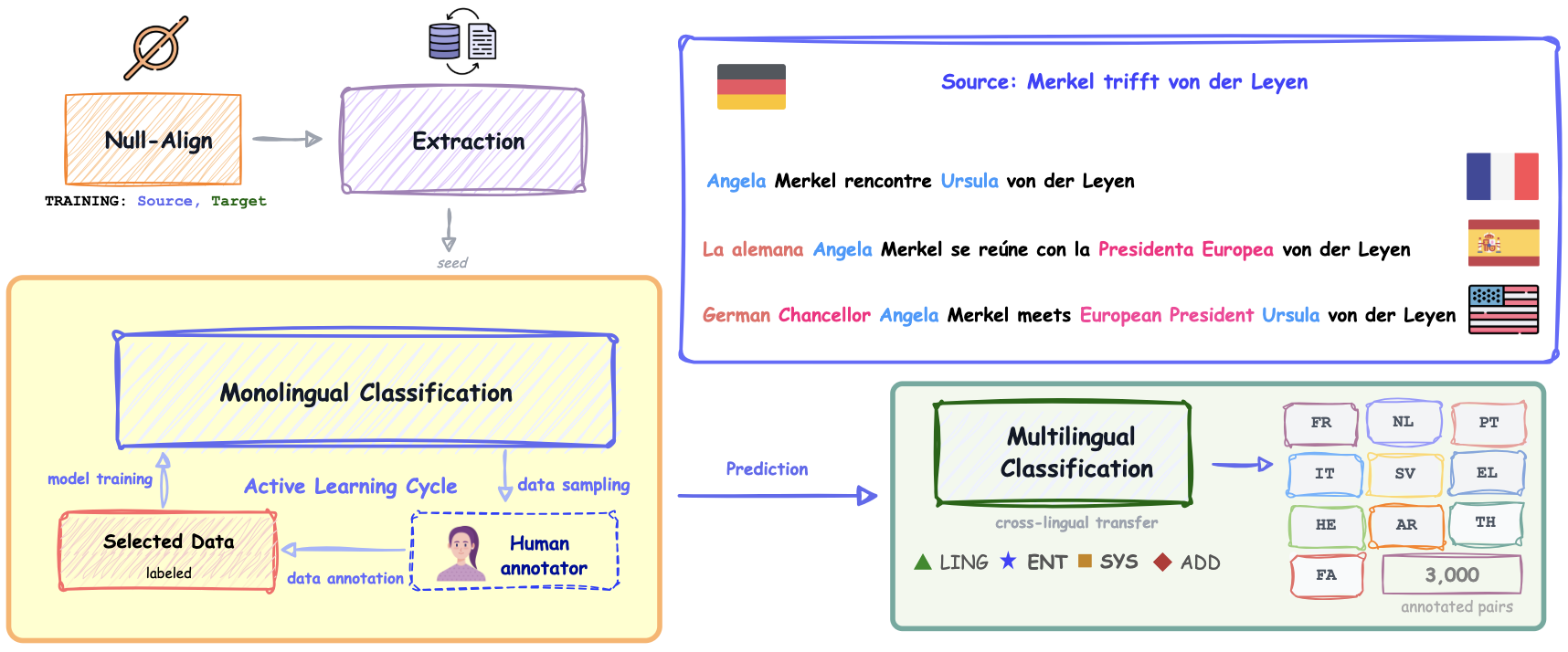}
    \caption{Main Findings. (Top left) Seed data is extracted from parallel corpora via null-alignments and refined through an active learning cycle with a human annotator, training a monolingual classifier. (Top right) Example sentence pairs illustrating pragmatic explicitation. (Bottom) The trained model transfers cross-lingually 
to predict explicitation labels across four broad categories— \emph{Entities} (\textcolor{blue}{$\bigstar$}\,\texttt{ENT}), \emph{System Transfers} (\textcolor{orange}{$\blacksquare$}\,\texttt{SYS}),
\emph{Linguistic Adjustments} (\textcolor{darkgreen}{$\blacktriangle$}\,\texttt{LING}), and
\emph{Added Information} (\textcolor{red}{$\blacklozenge$}\,\texttt{ADD}) across twelve 
language pairs, resulting in the PETra corpus of 3,000 annotated 
sentence pairs.}
    \label{fig:placeholder}
\end{figure*}

Most prior computational studies have focused on structural or discourse-level explicitations, such as the insertion of connectives or referential markers \citep{hoek2015role, lapshinova2017discovery}. However, pragmatic explicitation differs in that it targets implicit knowledge rather than linguistic form—it aims to make tacit, culture-dependent meaning explicit for a new audience. Despite its communicative and interpretive significance, no large-scale, multilingual, human-validated resource currently exists for systematically studying pragmatic explicitation as a measurable phenomenon.

This paper takes a step towards filling this gap. We present a \textbf{multilingual corpus of pragmatic explicitations}, systematically identified and annotated across two large parallel multilingual datasets: Europarl, and TED-Multi. 
We automatically extract candidate explicitations using \textit{null alignments}, i.e., words appearing only in one side of a translation, and refine them with named-entity recognition and part-of-speech constraints to target additions likely tied to cultural entities, roles, or institutions.

To ensure interpretive quality, we employ a human-in-the-loop annotation framework, combining expert annotation with active learning to iteratively improve coverage and precision. The resulting corpus, \textbf{PETra}, is the first multilingual, human-validated dataset for pragmatic explicitation detection.  \autoref{fig:placeholder} summarizes the overall pipeline.
Our contributions are as follows:
\begin{itemize}[itemsep=2pt, parsep=0pt, topsep=4pt]
    \item We formalize \textit{pragmatic explicitation} as a computationally tractable phenomenon and release PETra, the first multilingual, human-validated corpus for its study.
    \item We develop an active learning framework integrating linguistic heuristics with human feedback for efficient detection and annotation.
    \item We analyze explicitation patterns across 12 language pairs and two domains, uncovering systematic cross-linguistic tendencies in how translators encode cultural knowledge.
\end{itemize}

\section{Related Work}

Explicitation has long been a core concept in translation studies \citep{gellerstam1986translationese, baker1993corpus,laviosa1998universals, pym2005explaining, englund2005expertise}. Early comparative stylistics framed explicitation as a systematic shift from implicit to explicit expression across languages \citep{vinay1958stylistique}.
\citet{nida1964toward} framed similar additions as amplifications, emphasizing their role in improving readability or avoiding ambiguity derived from socio-cultural differences. \citet{blumkulka1986explicitation} conducted the first systematic empirical study, formulating the explicitation hypothesis, which broadly states that translations tend to be more explicit than non-translations. \citet{klaudy1993explicitation} has further distinguished obligatory, optional, pragmatic and translation-inherent explicitations, highlighting that some additions arise specifically to supply cultural or background knowledge.
Pragmatic explicitations are particularly important when translators anticipate that certain cultural or contextual information may be unknown to the target readers. Such explicitations go beyond syntactic or stylistic adjustments, aiming instead to bridge knowledge gaps and support effective intercultural communication \citep{snell2006turns}. While structural or obligatory explicitations are primarily driven by linguistic constraints, pragmatic explicitations are motivated by the communicative needs of the target audience.

Despite its centrality in translation studies, computational research on pragmatic explicitations remains limited. Early research emphasized discourse-level explicitations, such as the insertion of connectives or cohesive markers, as in the work of \citet{hoek2015role, lapshinova2017discovery,lapshinova2020coreference} or on general translationese patterns \citep{ teich2003cross,baroni2006new, volansky2015features, duttachowdhury-2020, chowdhury2021tracing}. \citet{kruger2020explicitation} investigated explicitation in machine-translated texts, though the focus was primarily on general explicitation phenomena rather than those arising from cultural or contextual asymmetries.
Recent research has shown growing interest in integrating cultural and contextual knowledge into multilingual systems. \citet{han-etal-2023-bridging} introduce WIKIEXPL, a dataset of concise explicative additions extracted from Wikipedia across languages, and demonstrate their utility in improving downstream tasks such as multilingual question answering. Their work highlights the practical benefits of targeted explicitations, though it focuses on generation for specific domains rather than creating a broadly validated corpus of pragmatic explicitations. \citet{lou2023audience} similarly proposed semi-automatic techniques for extracting explanations from aligned multilingual data, while \citet{yao2023empowering} curated culturally specific corpora to enhance cultural grounding in machine translation.

Automated detection of explicitations poses unique challenges: linguistic explicitations can often be identified via syntactic cues or cohesion markers, while pragmatic explicitations depend on world knowledge and cultural context. Furthermore, the rarity of pragmatic explicitations (often <1\% of tokens) makes annotation costly and sparse. Active learning has emerged as an effective strategy for handling such imbalances. 
Following recent advances in active learning for text classification \citep{wang-liu-2023-empirical, li2013active}, we adapt a multi-label active learning framework to pragmatic explicitation detection. The resulting corpus is the first systematic resource for studying this phenomenon computationally.

\section{Formalization of Cultural Explicitations}
We define cultural explicitations as fragments of translated text that
make implicit cultural or background knowledge explicit to the reader.
They go beyond literal or stylistic reformulation by adding information
that situates the text within the target audience’s cultural frame.
Below we outline the main conceptual dimensions; operational definitions are detailed in Section~\ref{annotation_design}.

\subsection{Conversion of Systems}
\label{sys_conv}
Texts are embedded in sociocultural systems—political, administrative,
educational, and measurement frameworks—that often have no direct
equivalents across languages. When translators adapt these constructs,
they convey cultural correspondences rather than lexical substitutions.

\paragraph{\textcolor{orange}{$\blacksquare$} Measurement systems.}
The best example of such cases is probably the differences between customary and metric systems which often require translation choices that add or replace information:
\begin{itemize}[nosep]
    \item replacement by conversion (e.g., ``1 mile'' → ``1.6km''; ``40 knots'' → ''75 km / h'')
    \item adding conversions (``120 miles'' → ``120 miles \emph{( 193 km )}'')
    \item adding a unit (``5 degrees'' → ``5 degrees \emph{C}'')
    \item adding dimension of a measurement (``8 cm'' → ``8 cm \emph{high}'')
    \item adding approximation terms (e.g., ``3 feet'' → ``\emph{about} one meter'')
    \item replacing figurative use (e.g., ``hundreds of miles'' → ``hundreds of kilometers'')
\end{itemize}
We exclude purely orthographic variants that do not alter meaning (e.g., ``lb'' → ``pound'', ; ''°'' → ``degree''; ``F'' → ``Fahrenheit'').

\paragraph{\textcolor{orange}{$\blacksquare$} Currencies.}
Additional cases for currencies include:
\begin{itemize}[nosep]
    \item adding a toponym (``\$'' → ``\emph{US} dollar''; ``105.000 pounds'' → ``105.000 \emph{British} pounds'')
    \item currency nicknames (``buck'' → ``dollar'')
\end{itemize}

\paragraph{\textcolor{orange}{$\blacksquare$} Education systems.}
Educational systems differ substantially across cultures, reflecting distinct institutional hierarchies, grading scales, and terminologies. Translating between education-related terms often requires interpreting underlying structural or conceptual differences, rather than performing a direct lexical substitution.  
Typical cases include:
\begin{itemize}[nosep]
    \item school names (e.g., ``college'' → ``Universität'')
    \item year names (e.g., ``sophomore'' →  ``second year of university'')
    \item grades and exam names (e.g., ``A'' → ``very good'')
    \item elite and support systems (e.g., ``Ivy League'' → ``universities of excellence'')
\end{itemize}

\paragraph{\textcolor{orange}{$\blacksquare$} Administrative bodies.}
Administrative bodies and political institutions reflect culture-specific governance structures. Transferring between these is not a pure translation, thus replacing or adapting such terms signals an attempt to orient the translation toward the target culture’s institutional framework.
\begin{itemize}[nosep]
    \item district namings (e.g., German ``Bundesland'' vs. US ``State'' and ``County'' vs. French ''Department'')
    \item traffic systems (e.g., ``highway'' vs. ``Autobahn''; ``subway'' vs. ``métro'')
    \item authorities and official institutions (e.g., US ``IRS'' vs. German ``Finanzamt''; US ``National Institute of Health'' vs. German ``Robert Koch Institut'')
\end{itemize}

These principles extend to other societal systems, including political parties, sports leagues, and public agencies, wherever system correspondence is culturally inferred.

\subsection{Named Entities}
Named entities are often culturally bound and require clarification or adaptation in translation. Cases include:
\paragraph{\textcolor{blue}{$\bigstar$} Replacing entity names.}
Widely known entities are often represented through acronyms (e.g., ``E.U.''). Culturally salient entities are often shortened to colloquial or generic nouns (e.g., ``the Wall'').
And sometimes entities are commonly referred to by other but similar entities (e.g., ``America'' for ``U.S.A.'').
The choice of abbreviation and replacement depend on the background knowledge within the respective culture. Abbreviations might be specific to the source culture (e.g., ``NIH'', ``the Mall''), ambiguous or misleading across cultures (``the Wall'' → ''the Berlin Wall'' vs. ``the Western Wall'') or in other cases more recognizable to the target audience (e.g., ``FBI''), in such cases replacement facilitates understanding.
Replacing entities conveys implicit cultural knowledge, highlighting the perceived equivalence between two entities and/or involve audience-oriented cultural mediation.
\begin{itemize}[nosep]
    \item replacing entity (e.g., ``United Kingdom'' → ``Great Britain'')
    \item colloquial forms (e.g., ``Aussie'' → ``Australian'')
    \item acronym expansion (e.g., ``NIH'' → ``National Institute for Health'')
    \item acronym collapsing (e.g., ``Federal Bureau of Investigation'' → ``FBI'')
    \item entity nouns (e.g., ``the Mall'' → ``the National Mall''; ``the States'' → ``the United States of America'')
\end{itemize}
\vspace{-1em}
\paragraph{\textcolor{blue}{$\bigstar$} Description and specifications.}
Entities are often reduced to their identifying core (e.g., ``Golden Gate''). Translators may reintroduce descriptive specifications to clarify meaning, as well as replace or augment a term with a functionally similar term of the target culture or descriptive explanations.
\begin{itemize}[nosep]
    \item hypernym (e.g., ``Golden Gate'' → ``Golden Gate \emph{Bridge}'')
    \item toponym (e.g., ``Cannery Row'' → ``Cannery Row \emph{(Californie)}'')
    \item ethnonym (``President Carter'' → ``\emph{American} President Carter'')
    \item matching target entity (e.g., ``FDA'' → ``American \emph{Ministry of Health}'')
    \item explanatory description (e.g., ``Ivy League'' → ``universities of excellence''; ``Mayor Bloomberg'' → ``the Mayor of New York''; ``EIA'' → ``\emph{US energy authority} EIA'')
\end{itemize}

\paragraph{Pure Translations.}
We explicitly exclude direct translations (e.g., ``Germany'' → ``Deutschland''), nominal component translations (e.g., ``Golden Gate Bridge'' → ``Puente Golden Gate'') and quotation marks, as these are pure lexical adaptions that do not add background information. 
Further, we exclude acronym expansion and acronym collapsing in cases where both entity and acronym are equally familiar across languages (e.g., ``U.N.'' → ``United Nations''; ``U.S.A.'' → ``United States''), as the choice between them reflects stylistic or linguistic conventions rather than cultural adaptation.

\subsection{Terminology}
\textcolor{darkgreen}{$\blacktriangle$} Some lexical items carry culturally loaded meanings or require disambiguation. Translators may add hypernyms (``hybrid'' → ``hybrid \emph{car}''), substitute hyponyms (``worker'' → ``commuter''), or expand acronyms (``AI'' → ``IA \emph{(intelligencia artificial)}''). We exclude simplifications (e.g., ``on the Earth'' → ``on the planet'') and idiomatic rephrasing.

\begin{table}[h!]
\scriptsize
    \centering
    \setlength{\tabcolsep}{3.5pt}
    \begin{tabular}{llp{3cm}}
    \toprule
    \textbf{Category} & \textbf{Subcategory} & \textbf{Description} \\
    \midrule
    \textcolor{blue}{$\bigstar$} ENT & ENT-REP & Entity Replacement \\
   & ENT-DESC & Entity Description \\
 & ENT-SPEC & Entity Specification \\
 & ENT-HYP & Entity Hypernym \\
 & ENT-ACR & Entity Acronym \\
    \midrule
    \textcolor{darkgreen}{$\blacktriangle$} LING & TRANS & Translation \\
 & LING-EXPL & Linguistic Explanation \\
  & ACR & Acronym \\
    & HYPER & Hypernym \\
   & HYPO-SPEC & Hyponym Specification \\
    \midrule
    \textcolor{orange}{$\blacksquare$} SYS & MEAS-CONV & Measurement Conversion \\
 & MEAS-DIM & Measurement Dimension\\
 & MEAS-SPEC & Measurement Specification\\
 & SYS-CONV & System Conversion \\
    & SYS-DESC & System Description \\
    \midrule
    \textcolor{red}{$\blacklozenge$} ADD & ADD-INF & Additional Information \\
 & CLEAR & Clarifying Information \\
 & DEIX & Deixis Resolution \\
    \bottomrule
    \end{tabular}
    \caption{Annotation Schema for Pragmatic Explicitation.
    Each main category is marked by a colored symbol: \textcolor{blue}{$\bigstar$}\,\textsc{ent}, \textcolor{orange}{$\blacksquare$}\,\textsc{sys}, \textcolor{darkgreen}{$\blacktriangle$}\,\textsc{ling}, \textcolor{red}{$\blacklozenge$}\,\textsc{add}.}
    \label{tab:annotation-schema-colored}
\end{table}

\subsection{Translator remarks}
\textcolor{red}{$\blacklozenge$} Translators occasionally insert short glosses: notes on wordplay (``sounds like sick brick''), contextual clarifications (``Adlai Stevenson'' → ``twice Democratic opponent of Eisenhower''), or literal translations (``Eat, Prey, Love'' → ``Prey = Beute''). Such remarks explicitly mediate cultural understanding. We exclude purely formal or stylistic changes that do not add cultural information; the operational categories are detailed in Section~\ref{annotation_design}.

\section{Corpus Construction and Annotation Framework}
We build a multilingual corpus of \textbf{pragmatic explicitations} by combining large-scale parallel data with automatic alignment heuristics and human-in-the-loop annotation. This section describes the extraction pipeline and annotation framework.

\subsection{Candidate Extraction via Null Alignments (\texttt{EXTR})}
We draw from two multilingual resources \textbf{Europarl} \citep{koehn2005europarl} and \textbf{TED-Multi} \citep{ye2018ted}.
We align each bilingual sentence pair forward-directionally using \texttt{eflomal} \citep{Ostling2016efmaral}\footnote{\url{https://github.com/robertostling/eflomal}} and \texttt{SimAlign} \citep{jalili-sabet-etal-2020-simalign}\footnote{\url{https://github.com/cisnlp/simalign}}. Tokens unaligned in the target are marked as \textit{additions}, potential cases of explicitation.

We use pre-trained spaCy pipelines to perform POS-tagging and NER, as well as compound decomposing in the case of German.
We extract sentence pairs containing both a named entity (PERSON, NORP, FAC, ORG, GPE, LOC, PRODUCT, EVENT, WORK\_OF\_ART, LAW, or LANGUAGE) and an unaligned content word (NOUN, PRON, or PROPN) in the target text, then annotate manually.

\subsection{Annotation Design}
\label{annotation_design}
For the Active Learning (AL), annotators classify candidates as:

\begin{itemize}
    \item \textbf{TRUE:} background or cultural enrichment.
    \item \textbf{FALSE:} direct or figurative translation without background or cultural enrichment.
    \item \textbf{DISCARD:} mistranslation, or non-aligned translations.
\end{itemize}

Annotation guidelines include positive/negative examples for each language. For the corpus, we annotate each detected explicitation with respect to its \textbf{type} (what kind of information is added or made explicit) and its \textbf{style} (how this information is integrated into the translation). \autoref{tab:annotation-schema-colored} provides an overview of the full annotation schema.

\subsubsection*{Type of Explicitations}
We distinguish four broad categories:
\emph{Entities} (\textcolor{blue}{$\bigstar$}\,\texttt{ENT}), \emph{System Transfers} (\textcolor{orange}{$\blacksquare$}\,\texttt{SYS}),
\emph{Linguistic Adjustments} (\textcolor{darkgreen}{$\blacktriangle$}\,\texttt{LING}), and
\emph{Added Information} (\textcolor{red}{$\blacklozenge$}\,\texttt{ADD}).
 
\paragraph{\emph{Entities} (\textcolor{blue}{$\bigstar$}\,\texttt{\textbf{ENT}})} This category covers modifications of named entities that make implicit cultural knowledge explicit.

\begin{compactdesc}
    \item \textbf{ENT-REP:} \emph{Entity replacement} — replacing an entity with another entity-term referring to the same referent (e.g., ``U.K.'' → ``Großbritannien'')
    
    \item \textbf{ENT-DESC:} \emph{Entity description} — describing what an entity is, instead of or in addition to matching it to a target-culture entity (e.g., ``EPA'' → ``die amerikanische Umweltschutzbehörde'').
    
    \item \textbf{ENT-SPEC:} \emph{Entity specification} — adding a toponym or other identifying element (e.g., ``Tucson'' → ``Tucson (Arizona)''; ``Library of Congress'' → ``La Bibliothèque du Congrès (USA)'').
    
    \item \textbf{ENT-HYP:} \emph{Entity hypernym} — adding a descriptive hypernym (e.g., ``Golden Gate'' → ``Golden Gate Brücke''; ``Oklahoma'' → ``das Musical Oklahoma'').
    
    \item \textbf{ENT-ACR:} \emph{Entity acronym} — adding, clarifying, or expanding an acronym (e.g., ``NYU'' → ``NYU (New York University)'') or an abbreviated form (e.g., ``the Mall'' → ``the National Mall'').
\end{compactdesc}

\paragraph{\emph{System Transfers} (\textcolor{orange}{$\blacksquare$}\,\texttt{\textbf{SYS}})} This category covers conversions between political, administrative, or measurement systems that differ across cultures.

\begin{compactdesc}
    \item \textbf{MEAS-CONV:} \emph{Measurement conversion} — conversion of a measurement from the system used in the source language culture to the system used in the target language culture (e.g., ``1 mile'' → ``1.6 km'').
    
    \item \textbf{MEAS-DIM:} \emph{Measurement dimension} — adding dimensional information to the measurement (e.g., ``3 feet'' → ``1m \emph{hoch}''; ``at 60 degrees'' → ``at \emph{a temperature of} 60 degrees'') 
    \item \textbf{MEAS-SPEC:} \emph{Measurement specification} —  specifying an implicit unit or scale (e.g., ``5 degree'' → ``5 degree \emph{Celsius}'')

    \item \textbf{SYS-CONV:} \emph{System conversion} — conversion of societal, political or administrative systems of the source language culture to the one of the target language culture (e.g., ``high school'' → ``Gymnasium'')
    
    \item \textbf{SYS-DESC:} \emph{System description} — explanatory transfers of system terms (e.g., ``freshman year'' → ``first year of university'') 
\end{compactdesc}

\paragraph{\emph{Linguistic Adjustments} (\textcolor{darkgreen}{$\blacktriangle$}\,\texttt{\textbf{LING}})} 
This category includes linguistic clarifications, terminological adjustments, and translator remarks that add explanatory information.

\begin{compactdesc}
    \item \textbf{TRANS:} \emph{Translation addition} — adding an additional translation of a term or phrase (e.g., ``Infinity Mushroom'' → ``Infinity Mushroom \emph{( Unendlichkeitspilz )}''). This also includes adding an original Latin spelling of a name in case of differing scripts.

    \item \textbf{LING-EXPL:} \emph{Linguistic explanation} —
    Remarks by the translator about the translation (e.g., ``loosely translated''; ``indirect speech''; ``transferred to our system''), linguistic remarks such as explanations of an idiom or pun (e.g., ``sounds like sick brick'').
    
     \item \textbf{HYPER:} \emph{Hypernym} — adding hypernyms (resp.~broader terms) to a noun, pronoun or entity, and/or other clarifying elements (e.g., ``hybrid'' → ``hybrid \emph{car}'')
     
    \item \textbf{HYPO-SPEC:} \emph{Hyponym specification} — replacing a term by a more specific term (e.g., ``worker'' → ``commuter''), or adding a specifying term to it (e.g., ``investments'' → ``\emph{health} investments'')

    \item \textbf{ACR:} \emph{Acronym} — adding an acronym or expansion of an acronym (e.g., ``AI'' → ``IA \emph{(intelligencia artificial)}'')
\end{compactdesc}

\paragraph{\emph{Added Information} (\textcolor{red}{$\blacklozenge$}\,\texttt{\textbf{ADD}})}  This residual category captures all remaining translator remarks that transport additional information.

\begin{table}[t]
\scriptsize
\centering
\setlength{\tabcolsep}{3.5pt}
\begin{tabular}{@{}l rrr rrrr r@{}}
\toprule
 & \multicolumn{3}{c}{Pairs} & \multicolumn{4}{c}{Types} & \\
\cmidrule(lr){2-4} \cmidrule(lr){5-8}
& \tiny{P} & \tiny{E} & \tiny{T}
 & \textcolor{blue}{$\bigstar$} & \textcolor{orange}{$\blacksquare$} & \textcolor{darkgreen}{$\blacktriangle$} & \textcolor{red}{$\blacklozenge$} 
 & $\Sigma$ \\
\midrule
\multicolumn{9}{@{}l}{\textit{\scriptsize TED-multi}} \\
DE & 534 & 105 & 162  & 146 & 708 & 177 & 41 & 1072 \\
ES & 372 &  47 &  63  & 161 & 352 &  49 &  8 & 570 \\
FR & 280 &  57 &  --  & 134 & 154 & 134 & 26 & 448 \\
PT & 211 &  55 &  --  &  85 &  104 &  87 & 33 & 309 \\
NL & 180 &  55 &  --  &  74 & 127 &  64 & 20 & 285 \\
IT & 123 &  47 &  --  &  56 &  84 &  49 & 16 & 205 \\
FA & 182 &  -- &  --  &  99 &  14 &  24 & 63 & 200 \\
HE & 109 &  -- &  --  &  29 &  35 &  29 & 30 & 123 \\
SV &  32 &  48 &  --  &  23 &  41 &  13 & 16 &  93 \\
EL &  13 &  59 &  --  &  38 &  22 &  13 &  9 &  82 \\
TH &  97 &  -- &  --  &  24 &  58 &  27 & 17 &  126 \\
AR &  65 &  -- &  --  &  23 &   10 &  30 & 17 &  80 \\
\midrule
\multicolumn{9}{@{}l}{\textit{\scriptsize Europarl}} \\
DE\textsubscript{ep} &  52 &  34 &  -- & 100 &  13 &  48 & 11 & 172 \\
\midrule
\textbf{Total} & & & & \textbf{992} & \textbf{1722} & \textbf{744} & \textbf{296} & \textbf{3754} \\
\bottomrule
\end{tabular}
\caption{Corpus statistics for PETra. Symbols match the type legend:
\textcolor{blue}{$\bigstar$}\,\textsc{ent},
\textcolor{orange}{$\blacksquare$}\,\textsc{sys},
\textcolor{darkgreen}{$\blacktriangle$}\,\textsc{ling},
\textcolor{red}{$\blacklozenge$}\,\textsc{add}. \textsc{P}=pool (classifier), \textsc{E}=extracted (null-alignment), \textsc{T}=active learning. Types are counted for individual instances of explicitation.}
\label{tab:corpus_stats}
\end{table}

\begin{table*}[t]
\scriptsize
\centering
\begin{tabular}{@{} p{1.2cm} p{4.2cm} p{8.2cm} c @{}}
\toprule
\textbf{Lang} & \textbf{Source (EN)} & \textbf{Target} & \textbf{}\\
\midrule
en\,$\to$\,de &
\textbf{[The EPA]} estimates, in the United States, by volume, this material occupies 25 percent of our landfills. &
\colorbox{cyan!15}{\strut Die amerikanische Umweltschutzbehörde}\E
schätzt, dass in den USA dieses Material, dem Volumen nach, 25\% unserer Mülldeponie ausmacht. &
\tiny{P} \\
\midrule

en\,$\to$\,de &
So, that's happening in the \textbf{[U.K.]} with \textbf{[U.K.]} government data. &
Das passiert also in \colorbox{cyan!15}{\strut Großbritannien}\E mit \colorbox{cyan!15}{\strut britischen}\E Regierungsdaten. &
\tiny{T} \\
\midrule

en\,$\to$\,de &
So basically, China is a SICK BRIC country. &
China ist also ein SICK-BRIC-Land. \colorbox{green!20}{\strut klingt wie ``kranker Ziegel''}\LI &
\tiny{P} \\
\midrule

en\,$\to$\,de &
And this bear swam out to that seal — \textbf{[800 lb.]} bearded seal — grabbed it, swam back and ate it. &
Und dieser Bär schwamm zu dieser Robbe hin — eine \colorbox{orange!20}{\strut 350 Kilo}\SY \colorbox{orange!20}{\strut schwere}\SY, bärtige Robbe — schnappte sie, schwamm zurück und aß sie. &
\tiny{P} \\
\midrule

en\,$\to$\,de &
You run up to \textbf{[15,000 feet]}, descend \textbf{[3,000 feet]}. &
Dann rennt man auf \colorbox{orange!20}{\strut eine Höhe von}\SY \colorbox{orange!20}{\strut 4.500 m}\SY hinauf, und wieder \colorbox{orange!20}{\strut 900 m}\SY bergabwärts. &
\tiny{E} \\
\midrule

en\,$\to$\,de &
It's a classic tale of Eat, Prey, Love. &
Die klassische Mär von `Eat, Prey, Love'. \colorbox{green!20}{\strut Prey = Beute}\LI &
\tiny{P} \\
\midrule

en\,$\to$\,nl &
So this is a Yellowstone, you know, of Saturn. &
Dit is een Yellowstone (\colorbox{cyan!15}{\strut natuurpark in de VS}\E \colorbox{red!15}{\strut bekend van de geisers}\AD), van Saturnus. &
\tiny{E} \\
\midrule

en\,$\to$\,nl &
You must have gotten your education \textbf{[here]}. &
U bent vast opgeleid \colorbox{red!15}{\strut in de VS}\AD \colorbox{red!15}{\strut en niet in India}\AD. &
\tiny{E} \\
\midrule

en\,$\to$\,es &
The \textbf{[seniors and juniors]} are driving the \textbf{[freshmen and the sophomores]}. &
Los \colorbox{orange!15}{\strut alumnos del último grado}\SY llevan a \colorbox{orange!15}{\strut los de primero}\SY. &
\tiny{P} \\
\midrule

en\,$\to$\,es &
They haven't learned that in Monterey. &
Eso no lo han aprendido en Monterey (\colorbox{cyan!15}{\strut California, EE.UU.}\E; \colorbox{red!15}{\strut sede de la conferencia}\AD). &
\tiny{P} \\
\bottomrule
\end{tabular}

\caption{Examples of pragmatic explicitation in TED-Multi. 
Colored symbols mark the category: 
\textcolor{blue}{$\bigstar$}\,\textsc{ent}~(entity), 
\textcolor{orange}{$\blacksquare$}\,\textsc{sys}~(system), 
\textcolor{darkgreen}{$\blacktriangle$}\,\textsc{ling}~(linguistic), 
\textcolor{red}{$\blacklozenge$}\,\textsc{add}~(added info). 
Source: \textsc{p}=pool (classifier), \textsc{e}=extracted (null-alignment), \textsc{t}=active learning.}
\label{tab:examples-pretty}
\end{table*}
\begin{compactdesc}
    \item \textbf{ADD-INF:} \emph{Additional information} — Adding additional (background) information that is not transported in the source text.
    Such as specifying time of certain events (e.g., ``on the 5th of September'' → ``Am 5.~September \emph{[ 2012 ]}''), description of what a word means (e.g., ``sarong'' → ``sarong \emph{( traditional Malaysian clothing )}''), additional information about an entity (e.g., ; ``Myanmar'' → ``Myanmar \emph{( former Birma )}''), and more.
    
    \item \textbf{CLEAR:} \emph{clarifying information} —
    Adding discourse-related context information, that is not contained in the source text, such as replacing a pronoun by its referee (e.g., ``go see him'' → ``go see him \emph{[ Dalai Lama ]}''). 
    This can be information known to the source audience, for example due to visual input, or prior discourse input.
    
    \item \textbf{DEIX:} \emph{deixis resolution} — resolving deixic references known to the audience (e.g. ``here'' → ``in the US'').
    
    \item \textbf{OTHER:} any other type of cultural explicitation.
\end{compactdesc}
\vspace{-0.7em}
\subsubsection*{Style of Explicitations} Each explicitation is additionally annotated for its integration style:
\begin{compactdesc}
    \item \textbf{R:} \emph{replace} -- replacing terms in the source by terms in the target (e.g., ``1 mile'' → ``\emph{1.6km}'').
    \item \textbf{A:} \emph{add} -- adding terms in the target additionally to the content of the source text (e.g., ``1 mile'' → ``1 mile \emph{(1.6km)}'').
\end{compactdesc}

\subsection*{Resulting Resource}
The resulting resource contains sentence pairs that show cultural explicitation in the translation. \autoref{tab:examples-pretty} shows some examples. Each pair contains:
\begin{compactdesc}
    \item \textbf{ID}: a unique identifier of the sentence pair, containing the language combination.

    \item \textbf{Source} / \textbf{Target}: the source language text and target language text. Brackets indicate parts of the text that were added or replaced by cultural explicitation in the translation.

    \item \textbf{Type}: indicated by colored symbols in the target text (see \S\ref{annotation_design}).

  \item \textbf{Source}: whether the pair was extracted via null-alignments (\textsc{extr}), detected by the classifier (\textsc{pool}), or annotated during active learning (\textsc{train}).
\end{compactdesc}

The final resource consists of instances extracted by null-alignment extraction \texttt{EXTR}, plus further instances found during active learning and instances detected by the final classifier \texttt{CLF}.

\autoref{tab:corpus_stats} summarizes the quantitative profile of the PETra corpus\footnote{Although more named-entity instances are extracted in Europarl, they are closely tied to the source text, often reflecting simple expansions or stylistic rephrasings. In contrast, TED translations exhibit more context-driven adaptation, with explicitations that provide additional background, explanations, or cultural information for the target audience.}.
The corpus covers 12 typologically diverse languages and includes both high-resource (e.g., English--German) and medium-resource pairs (e.g., English--Portuguese, English--Arabic), enabling cross-linguistic evaluation of explicitation behavior. Of all candidate sentence pairs, 6\% (2.6\% for TED, 42\% for Europarl\footnote{for TED on DE, ES, IT, PT, FR, NL, EL, SV, HE, AR, FA, TH and for Europarl on DE, ES, IT, PT, FR}) were extracted by \texttt{EXTR}.
3\% of the sentence pairs (2.5\% for TED, 7.7\% for Europarl) were classified as containing cultural explicitation by \texttt{CLF}.
Approximately 3,000 instances were manually labeled across the full dataset.

The types of explicitations vary across languages and domains. While on Europarl many explicitations concern specifications and hypernym additions to entities, TED shows a broader range, including linguistic explanations, descriptions and explanations of words and introduction of additional background information.
German and Spanish show a strong prevalence of system transfers, while Hebrew and Arabic for example prefer descriptions and translations of entities and concepts.

\section{Experiment Settings}
We adopt a multi-round, pool-based active learning framework inspired by \citet{wang-liu-2023-empirical}, tailored to multi-label explicitation classification. The process iteratively refines the labeled dataset through uncertainty-driven human feedback.
Our annotated seed set forms the initial training data for the active learner. Subsequent candidate samples for annotation are drawn automatically from the unlabeled pool using hybrid query strategies.

\subsection{Active Learning Workflow (\texttt{CLF})}
\paragraph{Data Splits.} 
We randomly sample a set of 400 annotated examples from the extracted data, and split it into 100 samples \textbf{seed data} and 300 samples \textbf{test data}. 
We ensure 1/3 of the seed data to be positive (explicitation) and 2/3 negative (non-explicitation) cases.
The test set remains fixed and unseen to the classifier throughout, serving only for final evaluation. The remaining samples serve as our training \textbf{pool}.
\vspace{-1em}
\paragraph{Model.}
We fine-tune \texttt{bert-base-multilingual-cased} for binary classification. Each instance consists of a sentence pair separated by the \texttt{[SEP]} token. Training uses the \texttt{Transformers} library with default settings for 10 epochs. 
A base classifier is trained on the seed set described above. 

\paragraph{Query Strategies.}
As pragmatic explicitations are sparse and diverse, we design a two-phase querying approach combining diversity-based exploration and uncertainty-based refinement. 
Each AL cycle queries 100 new instances, distributed across 2--3 of the strategies below following a heuristic sequence to balance diversity.

\paragraph{Combined Querying (8 iterations).}
We alternate between the following strategies:
\begin{itemize}
    \item \textbf{High-Confidence Positives:} selects high-probability positives based on lexical, POS, and alignment features, with random sampling above a fixed confidence threshold.
    
    \item \textbf{Embedding Clustering:} identifies unexplored regions via mini-batch $k$-means on sentence embeddings; one centroid per cluster is selected.

    \item \textbf{Diverse Seed Expansion:} expands around distant seed positives by retrieving diverse, nearby unlabeled instances in embedding space.
    
    \item \textbf{Nearest Positive Neighbors:} retrieves unlabeled instances closest (by cosine distance) to current positives; expanded to all positives over iterations.
    
    \item \textbf{Low-Confidence Sampling:} selects uncertain cases with lowest model confidence.
\end{itemize}

\paragraph{Uncertainty Sampling (5 iterations).}
After coverage stabilizes, we switch to uncertainty-driven sampling by selecting instances with posterior probabilities closest to 0.5 for the positive label.

\paragraph{Annotation and Iteration.}
Queried samples are manually annotated using the pragmatic explicitation schema (\S\ref{annotation_design}), merged with existing labels, and used to re-train the model.
We run 13 AL cycles in two phases. In the first phase (L1–L8), combined querying (100 samples per cycle) maximizes coverage of the explicitation space. We switch to uncertainty-only sampling for the second phase (L9–L13; 50 samples per cycle) once diversity-based strategies yield diminishing returns.
Finally, we construct an extended training set (L14) by adding 1,000 manually annotated instances from the German and Spanish pools,
as these provide a more diverse range of explicitation types (\autoref{tab:corpus_stats}),
strengthening generalization to rarer categories before cross-lingual transfer.

\section{Results and Discussion}

\begin{figure}
    \centering
    \begin{subfigure}[b]{0.48\textwidth}
        \centering
        \includegraphics[width=1\linewidth]{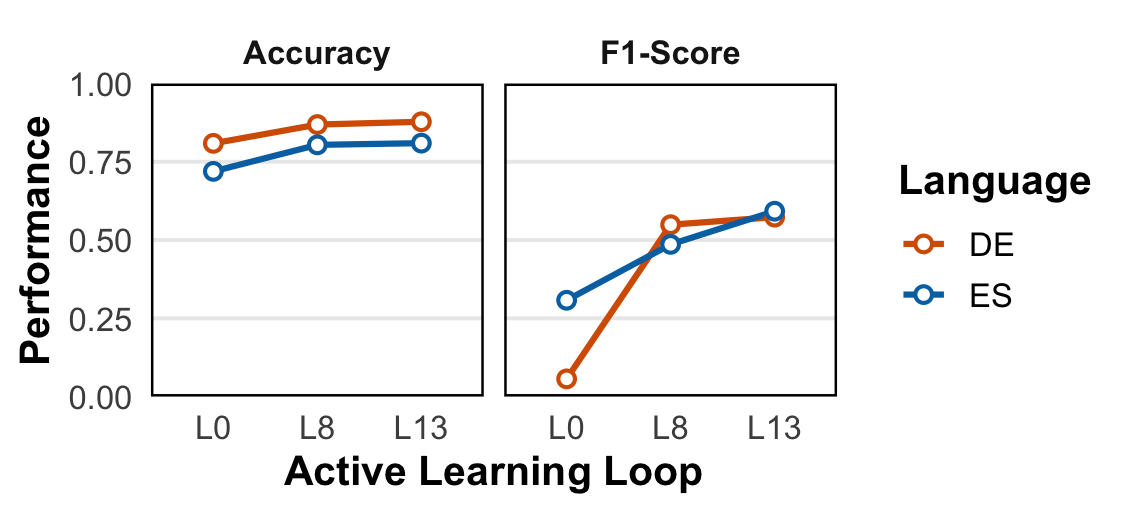}
        \caption{Monolingual Evaluation}
        \label{fig:classifier_performance}
    \end{subfigure}
    \begin{subfigure}[b]{0.48\textwidth}
        \centering
        \includegraphics[width=1\linewidth]{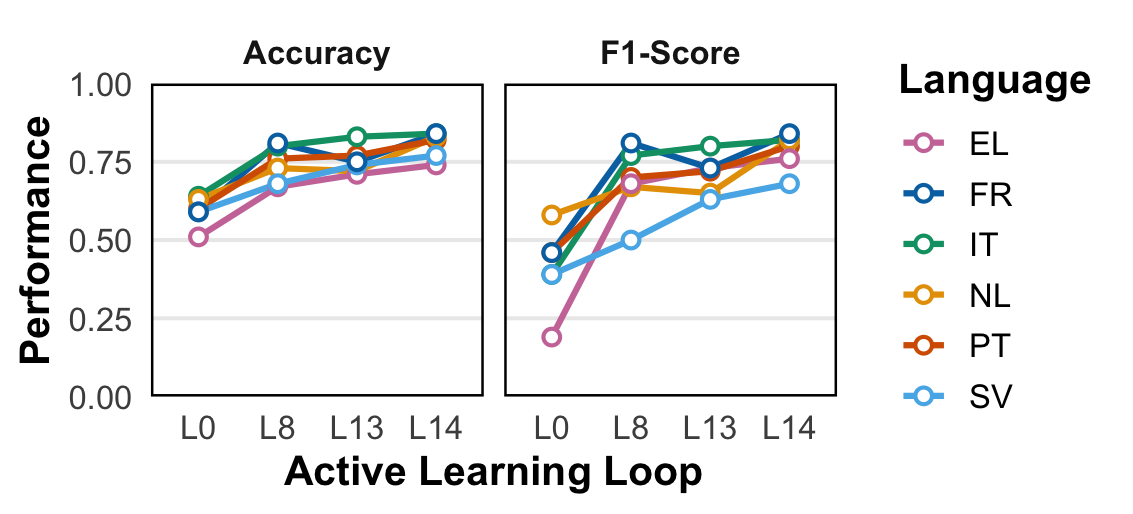}
        \caption{Cross-Lingual Evaluation}
        \label{fig:classifier_performance_Multi}
    \end{subfigure}
   \caption{Classifier accuracy and F1 at successive active learning stages (L0--L14), averaged over 5 random seeds. L0: before active learning, L8: after 8 cycles of combined strategies, L13: after additional 5 cycles of uncertainty-only strategies, 14: after adding additional 1000 samples of the German and Spanish pools.}
\end{figure}

We assess four versions of the classifier: trained on German and Spanish seed data only (L0), seed and data obtained from combined querying strategies (L8), seed and all data from active learning (L13), and seed and all data obtained during active learning plus additional 1000 annotated instances from German and Spanish pool (L14).
\paragraph{Monolingual Evaluation.}
\autoref{fig:classifier_performance} shows the performance of the Spanish and German classifiers at three stages of AL, as indicated on the x-axis. Active learning improves performance by roughly 7--8 percentage points, yielding approximately 1,150 high-confidence labeled examples for EN–DE and EN–ES. These examples are then used for large-scale automatic annotation of TED-Multi and Europarl, followed by manual spot checks.

\paragraph{Cross-Lingual Evaluation.}
We further train a cross-lingual classifier on the combined German and Spanish examples and evaluate it on 6 languages: Portuguese, Italian, French, Dutch, Swedish and Greek. The test sets for these languages each contain 100 annotated instances extracted via null-alignment methods\footnote{These datasets only contain very clear cases of explicitations, while the German and Spanish test sets contain also more subtle cases, due to the proficiency of the annotator.}.

\autoref{fig:classifier_performance_Multi} shows that cross-lingual transfer improves steadily across all languages. Accuracy and F1 scores are lowest at L0 and increase with each stage of active learning. The training languages, German and Spanish, start with relatively high scores and continue to improve, with German reaching 0.88 accuracy and 0.57 F1, and Spanish 0.81 accuracy and 0.59 F1 in L13. Languages close to the training data, such as Portuguese (similar to Spanish) and Dutch (similar to German), consistently achieve high scores, with L14 reaching 0.82–0.83 in accuracy and 0.80–0.82 in F1. Italian and French also perform well, showing strong cross-lingual transfer, while Swedish shows moderate gains. Greek, being more distant typologically, exhibits lower scores throughout, with accuracy peaking at 0.74 and F1 at 0.76 in L14.

\paragraph{Analysis.}
The results from the PETra experiments demonstrate that pragmatic explicitation is a consistent and measurable phenomenon across languages. \autoref{tab:corpus_stats} shows that entity-related and system level explicitations occur most frequently, reflecting translators' efforts to make culturally bound references and institutional systems more accessible to target readers. Linguistic adjustments (\textcolor{darkgreen}{$\blacktriangle$}\,LING) and additional-information cases (\textcolor{red}{$\blacklozenge$}\,ADD) appear less often but capture meaningful contextual enrichments, such as translator remarks and clarifying glosses.

Active learning proved essential for detecting these relatively rare phenomena. As illustrated in \autoref{fig:classifier_performance}, classifier performance improved by about seven to eight percentage points over successive annotation cycles, resulting in roughly 1150 high-confidence labeled examples for English–German and English–Spanish each. \autoref{fig:classifier_performance_Multi} further shows that classifier performance steadily improves across successive annotation cycles.
Two clear trends emerge: (i) performance improves consistently with more annotated data, and (ii) typological similarity to the seed languages strongly influences transfer success, with L14 achieving the best results across all languages.
\autoref{tab:examples-pretty} illustrates the qualitative range of the corpus, spanning entity descriptions, measurement conversions, system transfers, and translator glosses across three target languages. The diversity of examples confirms that PETra captures interpretable instances of culturally motivated explicitation.

\paragraph{Typological Patterns.}
While all languages in \textsc{PETra} exhibit a wide range of pragmatic explicitations, we observe systematic differences in how they occur across language families.
For instance, the category \textcolor{orange}{$\blacksquare$}\,\textsc{MEAS-DIM} (adding dimension information to measurements) is particularly common in Germanic languages compared to Romance and Semitic languages, showing that dimensional information may carry higher communicative salience for speakers of those languages.

European languages usually convert customary units to metric ones, while Semitic languages often keep the original measurement systems, reflecting broader cultural conventions around measurement and localization. 
On the other hand, German translations tend to assume more cultural background knowledge, using fewer explicitation strategies overall, while Semitic languages more often add descriptive explanations of concepts, entities, or linguistic forms (for example, explicitly stating that a term is adjectival, as in “Aristotelian science”). 

\section{Conclusion}
This paper introduces PETra, the first multilingual corpus and detection framework for pragmatic
explicitation. Through a combination of automatic extraction, active learning, and human annotation, we created a resource that captures how translators bridge world-knowledge gaps across languages and domains. The results demonstrate that pragmatic explicitation is both detectable and consistent across typologically diverse language pairs. Our active-learning framework improved classifier accuracy by approximately eight percentage points and achieved strong cross-lingual performance, confirming that such additions follow recognizable patterns of cultural mediation.

Future work will expand the dataset to additional languages, explore context-aware and multimodal signals, and investigate applications in machine translation and cultural adaptation modeling.
By documenting pragmatic explicitations at scale, PETra opens a new empirical direction for studying how translators make culture explicit, and represents a first step toward more culturally sensitive computational translation systems.

\section*{Acknowledgements}
Funded by the Deutsche Forschungsgemeinschaft (DFG, German Research Foundation) – SFB 1102 Information Density and Linguistic Encoding. CEB acknowledges her AI4S fellowship within the “Generación D” initiative by Red.es, Ministerio para la Transformación Digital y de la Función Pública, for talent attraction (C005/24-ED CV1), funded by NextGenerationEU through PRTR.

\section*{Limitations}
While PETra covers a diverse set of language pairs, including both European and non-European languages, it still reflects the availability of existing parallel corpora and therefore remains biased toward high-resource languages and formal domains such as TED talks and parliamentary proceedings. As a result, the observed explicitation patterns may not generalize to low-resource, oral, or literary contexts. 
The automatic extraction methods capture surface-level additions effectively but may overlook more implicit pragmatic or cultural adaptations. When used for analyzing multilingual large language models, the dataset can reveal systematic differences in cultural representation, yet it cannot isolate the underlying causes of such differences (e.g., training data imbalance or alignment procedures). Consequently, results should be interpreted as diagnostic indicators rather than exhaustive measures of cross-cultural representation.

\section*{Bibliographical References}\label{sec:reference}

\bibliographystyle{lrec2026-natbib}
\bibliography{lrec2026-example}

\end{document}